\documentclass[letterpaper, 10 pt, conference]{ieeeconf}  

\IEEEoverridecommandlockouts                              

\usepackage{hyperref}

\hypersetup{
  pdfauthor={Renaud Dube'},
  pdftitle={SegMatch: Segment Based Place Recognition in 3D Point Clouds},
  pdfsubject={SegMatch: Segment Based Place Recognition in 3D Point Clouds},
  urlcolor=blue,
}

\usepackage[numbers,sort&compress]{natbib}

\usepackage{graphicx} 
\usepackage{epsfig} 
\usepackage{verbatim} 
\usepackage{color}
\usepackage{soul}
\usepackage{subfig}
\usepackage[font=footnotesize]{caption}

\usepackage{amssymb}
\usepackage{amsmath}
\usepackage{algorithm}
\usepackage{algpseudocode}

\usepackage[english]{babel}
\usepackage[T1]{fontenc}
\usepackage[utf8]{inputenc}

\usepackage{xspace}

\usepackage{todonotes}

\usepackage{epstopdf}

\usepackage{physics}

\usepackage[nolist,nohyperlinks]{acronym}
\usepackage{verbatimbox}

\begin{document}

\begin{minipage}{0.7\paperwidth}
\begin{center}
This paper has been accepted for publication in IEEE International Conference on Robotics and Automation.

\vspace{5mm}
DOI: 10.1109/ICRA.2017.7989618

\vspace{10mm}
Please cite our work as:

\vspace{5mm}
R. Dub\'e, D. Dugas, E. Stumm, J. Nieto, R. Siegwart, and C. Cadena. ``SegMatch: Segment Based Place Recognition in 3D Point Clouds.'' IEEE International Conference on Robotics and Automation (ICRA), 2017.
\end{center}

\vspace{10mm}
bibtex:
\begin{verbnobox}[\small]
@inproceedings{segmatch2017,
  title      =  {{SegMatch}: Segment Based Place Recognition in 3D Point Clouds},
  author     =  {Dub{\'e}, Renaud and Dugas, Daniel and Stumm, Elena and
                Nieto, Juan and Siegwart, Roland and Cadena, Cesar},
  booktitle  =  {IEEE International Conference on Robotics and Automation (ICRA)},
  year       =  {2017}
}
\end{verbnobox}
\end{minipage}

\title{\LARGE \bf
\textit{SegMatch}: Segment Based Place Recognition in 3D Point Clouds
}

\author{Renaud Dub\'e \and Daniel Dugas \and Elena Stumm \and Juan Nieto \and Roland Siegwart \and Cesar Cadena$^{*}$
\thanks{$^{*}$Authors are with the Autonomous Systems Lab, ETH, Zurich
        {\tt\small authors@mavt.ethz.ch}.}%
        \thanks{
This work was supported by the European Union's Seventh Framework Programme for Research and Technological Development under the TRADR project No. FP7-ICT-609763.
}
}

\maketitle
\thispagestyle{empty}
\pagestyle{empty}

\begin{abstract}
%
%
%
Place recognition in 3D data is a challenging task that has been commonly approached by adapting image-based solutions.
Methods based on local features suffer from ambiguity and from robustness to environment changes while methods based on global features are viewpoint dependent.
We propose \textit{SegMatch}, a reliable place recognition algorithm based on the matching of 3D segments.
Segments provide a good compromise between local and global descriptions, incorporating their strengths while reducing their individual drawbacks.
\textit{SegMatch} does not rely on assumptions of `perfect segmentation', or on the existence of `objects' in the environment, which allows for reliable execution on large scale, unstructured environments.
We quantitatively demonstrate that \textit{SegMatch} can achieve accurate localization at a frequency of 1Hz on the largest sequence of the KITTI odometry dataset.
We furthermore show how this algorithm can reliably detect and close loops in real-time, during online operation. 
In addition, the source code for the \textit{SegMatch} algorithm is made publicly available\footnote{The \textit{SegMatch} algorithm is available at \url{https://github.com/ethz-asl/segmatch} and a video demonstration is available at \url{https://www.youtube.com/watch?v=iddCgYbgpjE}.}.

%
%

\end{abstract}

\section{INTRODUCTION}
Place recognition represents one of the key challenges of accurate Simultaneous Localization and Mapping (SLAM).
As drift is inevitable when performing state estimation without global positioning information, reliable loop-closure detection is a crucial capability for many robotic platforms \cite{thrun2002robotic}.
%
%
%
Many successful strategies for performing place recognition using images are proposed in the literature.
%
%
However, image-based place recognition can become unreliable when strong changes in illumination occur, and in the presence of strong viewpoint variations \cite{lowryvisual}.
%
%
Lidar-based localization, on the other hand, does not suffer from changes in external illumination, and since it captures geometry in a very fine resolution, does not suffer as much as vision when changes in viewpoint are present.
This paper therefore considers 3D laser range-finders for their potential to provide robust localization in outdoor environments.

Current strategies for recognizing places in 3D laser data are primarily based on keypoint detection and matching \cite{bosse2013place}.
%
%
%
In the context of performing place recognition on images, \citet{lowryvisual} state that using descriptors at the level of segments or objects could provide the benefits of both local and global feature approaches.
Object or segment maps also offer several advantages over their metric and topological counterparts.
Among others, these maps better represent situations where static objects can become dynamic, and are more closely related to the way humans perceive the environment \cite{thrun2002robotic}.

\begin{figure}
\centering
\includegraphics[width=3.4in]{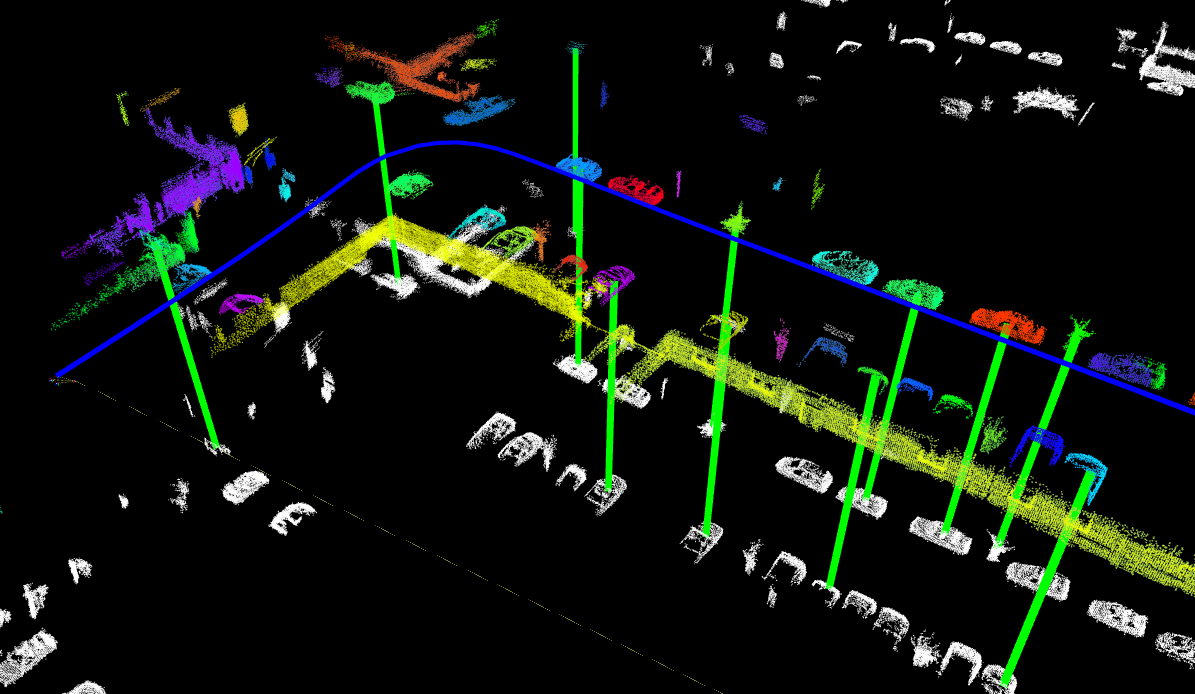}
\caption{An illustration of the presented place recognition framework. The reference point cloud is shown below (in white), and the local point cloud is aligned above. Colours are used to show the point cloud segmentation, and segment matches are indicated with green lines.}
\label{fig_lc7_cut}
\vspace{-5mm}
\end{figure}

While working at the level of objects would be ideal, it also has a twofold assumption.
First, that we have access to a perfect object segmentation technique, and second, that there are actual `objects' in the environment, under the definition of \cite{alexe2010cvpr}.
These assumptions do not hold in general because of imperfect segmentation and because of common real-world scenarios with no distinguishable objects.
%
%
This work therefore introduces \textit{SegMatch}, a segment-based approach which takes advantage of more descriptive shapes than keypoint-based features without the aforementioned strong assumptions of object-based approaches.
In other words, we recognize places by matching segments that belong to partial or full objects, or to parts of larger structures (windows, arcs, façades).
Examples of such segments can be seen in Fig.~\ref{fig_lc7_cut} for data collected in an urban scenario.


Our system presents a modular design. It first extracts and describes segments from a 3D point cloud, matches them to segments from already visited places and uses a geometric-verification step to propose place recognition candidates.
One advantage of this segment-based technique is its ability to considerably compress the point cloud into a set of distinct and discriminative elements for place recognition.
%
We show that this does not only reduce the time needed for matching, but also decreases the likelihood of obtaining false matches.

When it comes to segment description, although numerous 3D point cloud descriptors exist \cite{scovanner20073,wohlkinger2011ensemble,rusu2009fast}, there is no clear evidence of relative performance among them, such as power of generalization or robustness against symmetry in geometry for instances.
%
Therefore, we have opted for a machine learning approach to match a variety of standard descriptors computed over the segments.
Nonetheless, due to the modular nature of the presented framework, future advances in 3D segmentation, recognition, and description can be used by replacing the respective components in our pipeline.
%
%

To the best of our knowledge, this is the first paper to present a real-time algorithm for performing loop-closure detection and localization in 3D laser data on the basis of segments.
%
%
%
More specifically, this paper presents the following contributions:

\begin{itemize}
\item \textit{SegMatch}, a segment based algorithm to perform place recognition in 3D point clouds.
\item An open source implementation of \textit{SegMatch} for online, real-time loop-closure detection and localization.
\item A thorough evaluation of the algorithm performances in real-world applications.
\end{itemize}

The paper is structured as follows: Section~\ref{sec:related_work} provides an overview of the related work in the field of place recognition in 3D point clouds.
The proposed algorithm is then described in Section~\ref{sec:system} and evaluated in Section~\ref{sec:experiments}.
Section~\ref{sec:conclusion} finally concludes with a short discussion.

\section{RELATED WORK}
\label{sec:related_work}
Detecting loop-closures from 3D data is still an open problem in robot localization.
The problem has been tackled with different approaches.
We have identified three main trends: (i) approaches based on local features, (ii) global descriptors and (iii) based on planes or objects.

The works presented in \cite{bosse2013place, zhuang20133, steder2010robust, steder2011place, Gawel2016} propose to extract local features from keypoints and perform matching on the basis of these features.
\citet{bosse2013place} extract keypoints directly from the point clouds and describe them with a 3D \textit{Gestalt} descriptor.
Keypoints then vote for their nearest neighbors in a \textit{vote matrix} which is finally thresholded for recognizing places.
A similar approach has been used in \cite{Gawel2016}.
Apart from such Gestalt descriptors, a number of alternative local feature descriptors exist which can be used in similar frameworks.
This includes features such as Fast Point Feature Histogram (FPFH)~\cite{rusu2009fast} which will also be employed later in this work.
%
Alternatively, \citet{zhuang20133} transform the local scans into bearing-angle images and extract Speeded Up Robust Features (SURFs) from these images.
A strategy based on 3D spatial information is employed to order the scenes before matching the descriptors.
A similar technique by \citet{steder2010robust} first transforms the local scans into a range image.
Local features are extracted and compared to the ones stored in a database, employing the Euclidean distance for matching keypoints.
This work is extended in \cite{steder2011place} by using Normal-Aligned Radial Features (NARF) descriptors and a bag of words approach for matching.
%
%
%
\citet{zhang2014loam} are able to estimate odometry in real-time using range data.
Loop-closures are mentioned but rely on an offline algorithm.
%




Using global descriptors of the local point cloud for place recognition is also proposed \cite{rohling2015fast,granstrom2011learning,magnusson2009automatic}.
\citet{rohling2015fast} propose to describe each local point cloud with a 1D histogram of point heights, assuming that the sensor keeps a constant height above the ground.
%
%
The histograms are then compared using the \textit{Wasserstein} metric for recognizing places.
\citet{granstrom2011learning} describe point clouds with rotation invariant features such as volume, nominal range, and range histogram.
Distances are computed for scalar features and cross-correlation for histogram features, and an AdaBoost classifier is trained to match places.
Finally, \ac{ICP} is used for computing the relative pose between point clouds.
In another approach, \citet{magnusson2009automatic} split the cloud into overlapping grids and compute shape properties (spherical, linear, and several type of planar) of each cell and combine them into a matrix of surface shape histograms.
%
%
Similar to other works, these descriptors are compared for recognizing places.
%


While local keypoint features often lack descriptive power, global descriptors can struggle with invariance.
Therefore other works have also proposed to use 3D shapes or objects for the place recognition task.
\citet{fernandez2013fast}, for example, propose to perform place recognition by detecting planes in 3D environments.
%
%
%
The planes are accumulated in a graph and an interpretation tree is used to match sub-graphs.
A final geometric consistency test is conducted over the planes in the matched sub-graphs.
%
%
The work is extended in \cite{fernandez2016scene} to use the covariance of the plane parameters instead of the number of points in planes for matching.
This strategy is only applied to small, indoor environments and assumes a plane model which is no longer valid in unstructured environment.
A somewhat analogous, seminal work on object-based loop-closure detection in indoor environments using RGB-D cameras is presented by \citet{finman2015icraws}.
Although presenting interesting ideas, their work can only handle a small number of well segmented objects in small scale environments.

We therefore aim for an approach which does not rely on assumptions about the environment being composed of simplistic geometric primitives such as planes, or a rich library of objects.
This allows for a more general, scalable solution.
Inspiration is taken from \citet{douillard2012scan} and \citet{nieto2006scan} which proposed different SLAM techniques based on segments.
A strategy for aligning Velodyne scans based on segments is proposed in \cite{douillard2012scan} where the \textit{Symmetric Shape Distance} is used to compare and match segments as defined in \cite{douillard2014pipeline}.
Analogously, \cite{nieto2006scan} proposed an Extended Kalman Filter solution which uses segments as landmarks, rather than point features.

\section{\textit{SegMatch} Algorithm}
\label{sec:system}
\begin{figure*}
\centering
\includegraphics[width=4.5in]{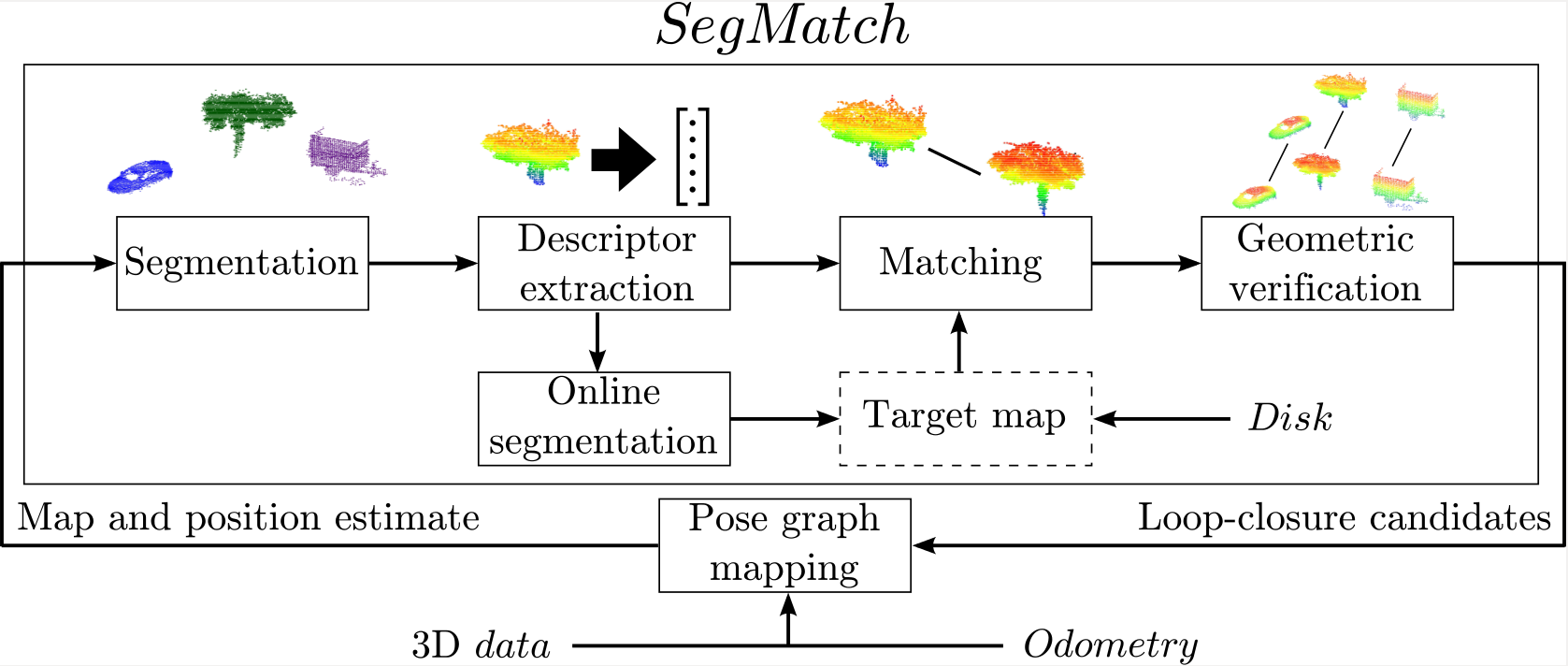}
\caption{Block diagram of \textit{SegMatch}, a modular place recognition algorithm. The target map can either be loaded from disk (for localization) or computed online (for loop-closure).}
\label{fig_schema_bloc}
\vspace{-5mm}
\end{figure*}

In this section we describe our approach for place recognition in 3D point clouds.
The proposed system is depicted in Fig.~\ref{fig_schema_bloc} and is composed of four different modules: point cloud segmentation, feature extraction, segment matching, and geometric verification.
Modularity has been a driving factor during the design phase.
In the following, we propose an example implementation for every module of the system which can easily be adjusted for operating in different types of environment.
%


\subsection{Segmentation}
\label{sec_feature_extraction}

The first building block of \textit{SegMatch} segments point clouds into distinct elements for matching.
For each incoming point cloud $P$ given in a global reference frame, we first extract a local point cloud by defining a cylindrical neighbourhood of radius $R$, centred around the current robot location.
A voxel grid is then applied to the resulting source cloud in order to filter-out noise in voxels where there is not enough evidence for occupancy.
The filtered point cloud is finally segmented into a set of point clusters $C_i$ using the "Cluster-All Method" of \cite{douillard2011segmentation}.
This segmentation requires the ground plane to be previously removed, which can be achieved by clustering adjacent voxels based on vertical means and variances \cite{douillard2011segmentation}.
Once the ground plane is removed, Euclidean clustering is used for growing segments.
For each cluster $C_i$ the centroid $c_i$ is computed as the average of all its points. 

%

\subsection{Feature extraction}
\label{sec_feature_extraction}
Once we have segmented the point-cloud, we extract features for each segment.
This feature extraction step is used for compressing the raw data and builds segment signatures suitable for recognition and classification.
As there is no clear \emph{gold-standard} descriptor for 3D data, we use several different descriptors.

Given a cluster $C_i$, descriptors are computed resulting in feature vector $f_i = \begin{bmatrix} f_i^1 & f_i^2 & \ldots & f_i^m \end{bmatrix}$.
Whereas this feature vector could be extended to include a large quantity of descriptors, two descriptors which produced interesting results are here presented.

$f^1$ \textit{Eigenvalue based}: In this descriptor, the eigenvalues of the segment's point cloud are computed and combined in a feature vector of dimension 1x7.
We compute the \textit{linearity}, \textit{planarity}, \textit{scattering}, \textit{omnivariance}, \textit{anisotropy}, \textit{eigenentropy} and \textit{change of curvature} measures as proposed in \cite{weinmann2014semantic}.

$f^2$ \textit{Ensemble of shape histograms}: This feature of dimension 1x640 is made of 10 histograms which encode the shape functions D2, D3 and A3 as described in  \cite{wohlkinger2011ensemble}.
The D2 shape function is a histogram of the distances between randomly selected point pairs while D3 encodes the area between randomly selected point triplets.
The A3 shape function describes the angles between two lines which are obtained from these triplets.

\subsection{Segment matching}

Using these features, we wish to identify matches between segments from the source and target clouds.
For this operation we opted for a learning approach, as it is often difficult to select the appropriate distance metric and thresholds, especially when multiple feature types are involved.
A classifier is therefore used to make the final decision about whether two segments represent the same object or object parts.
In order to maintain efficiency, we first retrieve candidate matches by performing a k-d tree search in the feature space, which are then fed to the classifier.

Specifically, we employ a random forest for its classification and timing performances.
The idea behind this classifier is to construct a multitude of distinct decision trees and to have them vote for the winning class.
During the learning phase, each tree is trained using a bootstrapped subset of the training data set and a random subset of features.
Random forests offer classification performance similar to the AdaBoost algorithm but are less sensitive to noise in the output label (such as a mis-labeled candidates) since they do not concentrate their efforts on misclassified candidates \cite{breiman2001random}.
Random forests can also provide information regarding the feature's relative importance for the classification task.

For the random forest classifier to determine whether clusters $C_i$ and $C_j$ represent the same object, we compute the absolute difference between the eigenvalue based feature vectors: $\Delta f^1 = \abs{f_i^1 - f_j^1}$.
The feature vectors $f_i^1$ and $f_j^1$ are also fed to the classifier for a total eigenvalue based feature dimension of 1x21.
For the ten histograms of the ensemble of shape features, the histogram intersection is computed, resulting in a feature of dimension 1x10.
Given this set of features, the random forest classifier assigns a classification score $w$ of being a match.
A threshold on $w$ is applied for building the final list of candidate matches passed to the next module.
This threshold value is dependent on the subset of features used for matching and is defined in Table~\ref{tab_parameters}.

\subsection{Geometric verification}
\label{sec:geometric_verification}

The candidate matches are fed to a geometric-verification test using random sample consensus (RANSAC) \cite{fischler1981random}.
Transformations are evaluated using the segment centroids.
%
%
A geometrically-consistent cluster of segments is finally accepted based on a minimum number of segments in it, resulting in a 6DOF transformation and a list of matching segments which represent a place recognition candidate.

\section{EXPERIMENTS}
\label{sec:experiments}

The proposed segment based  algorithm is evaluated using the KITTI odometry dataset \cite{geiger2012we}.
We first illustrate how this dataset can be processed for generating segment matching samples for training and testing the classifiers (Sections~\ref{ssec:dataset} and ~\ref{ssec:train_test}).
This leads to an analysis of the performances of different classifiers' parametrization (Section~\ref{ssec:class_perf}).
The segment based localization strategies are then compared to a keypoint approach as a baseline (Section~\ref{ssec:localization_performance}).
We then show how the segment based loop detection framework can be used for online place recognition applications and how it can successfully operate in different environments (Sections~\ref{ssec:loop_closure_performance} and ~\ref{ssec:clausius}).



\subsection{Dataset}
\label{ssec:dataset}

The following three analyses are performed using sequences 00, 05 and 06 of the KITTI  dataset.
Sequence 06 lasts 1.2~km (114 seconds) and is only used for training the classifiers.
%
%
Sequence 00 lasts 3.7~km (470 seconds) and is particularly interesting as it contains one large loop where the vehicle revisits the same environment for a stretch of 500 meters.
This section with multiple traversals will therefore be used in the localization experiment.
Sequence 05 lasts 2.2~km (287 seconds) and is used for presenting the online operation of the framework.

As previously described, the input of our segment based place recognition algorithm is a point cloud in a global reference frame.
For generating a point cloud in real-time from the large quantity of measurements provided by a Velodyne sensor, a uniform rate sub-sampling filter is first applied for removing half of the scan's points.
These scans are added to the point cloud every time the robot drove a minimum distance of 1 meter.
As the sensor configuration on-board the car is known (e.g. height of the sensor), ground plane extraction is performed by directly filtering the input scan by minimum height.
This simple assumption is efficient and works very well for this driving dataset.

For extracting the source point cloud, the radius of the cylindrical neighbourhood $R$ is set to 60 meters.
The voxel grid leaf size is set to 0.1 meters, and a minimum of two points within a voxel is required to consider it as occupied.
For segmentation, the maximum Euclidean distance between two occupied voxels such that they are considered to belong to the same cluster is set to 0.2 meters.
We choose to consider only segments which contain a minimum of 100 points and a maximum of 15000 points.

\subsection{Training and testing setup}
\label{ssec:train_test}

The following procedure is performed for generating both training and testing data.
During the first section of a given sequence, a target map is generated and processed by extracting and describing segments.
When the vehicle revisits the same section of the environment, the ground truth information is used for storing pairs of corresponding and differing segments from the source and target clouds.
For each segment in the local cloud, we perform \ac{k-NN} retrieval in feature space and identify the 200 nearest neighbors in the target map.
These candidates are saved as true matches for the corresponding segments and false matches for differing segments.
Using this procedure on sequence 06 of the KITTI  dataset, we generate 2000 true and 800000 false segment matches.
For training the random forests, we adopt a 1:50 ratio between the number of positive and negative samples which results in a training set of 102000 samples.


\begin{table}
  \renewcommand{\arraystretch}{1.3}
  \centering
  \caption{Parameters of three segment matching strategies.}
  \label{tab_parameters}
  \resizebox{3.3in}{!}{%
  \begin{tabular}{|c|c|c|c|}

    \hline
    Parameter & \textit{L2} & \textit{rf\_eigen} & \textit{rf\_full+shapes} \\
    \hline
    Number of neighbors & \multicolumn{3}{c|}{200} \\
    k-NN Feature space & \multicolumn{3}{c|}{Eigenvalue based}\\
    Hard threshold value & 0.0024 & N/A & N/A \\
    Number of trees & N/A & 25 & 25\\
    Threshold on probability & N/A & 0.81 & 0.72 \\
    Minimum cluster size & \multicolumn{3}{c|}{4} \\
    RANSAC resolution & \multicolumn{3}{c|}{0.4 meter} \\
    \hline
  \end{tabular}}
  \vspace{-4mm}
\end{table}

\subsection{Segment matching performance}
\label{ssec:class_perf}

The goal of this first experiment is to evaluate the performances of three segment matching techniques.
The first strategy titled \textit{L2} applies a threshold on the Euclidean distance between two segment's features vectors.
The second strategy, \textit{RF\_eigen}, is based on a random forest which relies only on the eigenvalue based features.
The last strategy, \textit{RF\_eigen+shapes}, uses the full set of features described in Section~\ref{sec_feature_extraction}.
The parameters used for each classifier are summarized in Table~\ref{tab_parameters}.



\begin{figure}
\centering
\includegraphics[width=2.5in]{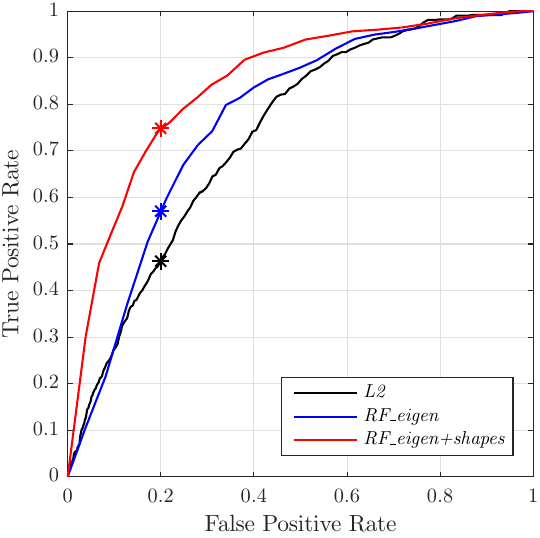}
\caption{ROC curves for segment matching performance using a hard threshold on the distance between segment features (\textit{L2}) compared to using random forests on two different feature sets (\textit{RF\_eigen} and \textit{RF\_eigen+shapes}). The operating points of FPR = 0.2 are indicated. 
}
\label{fig_roc_curves}
\end{figure}

\begin{figure*}[!t]
\centering
\fbox{\includegraphics[width=6.8in]{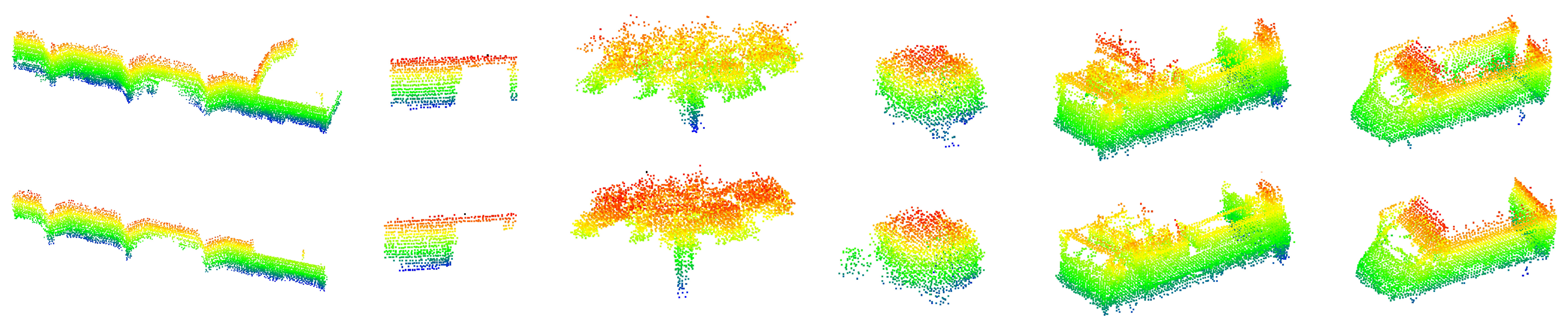}}
  \caption{Corresponding segments successfully detected by the \textit{SegMatch} algorithm. The top and bottom rows illustrate segments from the target and the source clouds respectively.}
  \label{fig:segments}
\end{figure*}

Fig.~\ref{fig_roc_curves} shows the receiver operating characteristic (ROC) curves of the three methods when testing on data extracted from sequence 00.
The random forest classifiers offer an improvement in performance when compared to their L2 norm counterpart.
Examples of corresponding segments correctly identified by the \textit{RF\_eigen+shapes} strategy are illustrated in Fig. ~\ref{fig:segments}.

In the experiments of the following section, we illustrate how these classifiers perform during real-time localization.
We define a false positive rate (FPR) of 0.2 to be the operating point of all classifiers in order to limit false segment matches and avoid false place recognitions.
This parameter and the other ones summarized in Table~\ref{tab_parameters} are used for the localization and loop-closure experiments of Sections~\ref{ssec:localization_performance} and~\ref{ssec:loop_closure_performance}.


\subsection{Localization performance}
\label{ssec:localization_performance}

This section evaluates the performance of the \textit{SegMatch} algorithm for localizing in a target segment map.
The section of interest in sequence 00 (as described in Section~\ref{ssec:dataset}) is used for creating the target map, and localization is performed when this section is revisited.
The three segment based strategies described in section ~\ref{ssec:class_perf} are compared to a keypoint based place recognition technique.

%
%

\subsubsection{Keypoint baseline}
For the \textit{keypoint} localization method, normals are first computed for every point of the filtered cloud using a radius of 0.3 meters.
Keypoints are selected in the target and source clouds using the Harris 3D keypoint extractor of the PCL library \cite{Rusu_ICRA2011_PCL}.
These keypoints are filtered to have a minimum distance of 0.5 meters between each keypoint.
This ensures that the same regions are not described twice, which in turn reduces ambiguity during the later geometrical verification step.
Each keypoint is described using the Fast Point Feature Histogram (FPFH) with a radius of 0.4 meters \cite{rusu2009fast}.
The source keypoints are matched to their 75 closest neighbors in the target cloud and the geometric verification algorithm described in Section~\ref{sec:geometric_verification} is used to filter this list of keypoint matches and to output localizations.
Parameters were chosen in an attempt to get the best performance we could find.

\subsubsection{Results}
\begin{figure}
\centering
\includegraphics[width=2.8in]{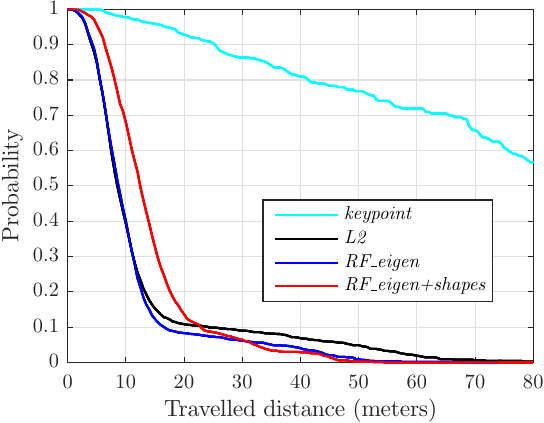}
\caption{Probability of travelling a given distance before localizing in the target segment map. Data is obtained from 90 localization runs for each strategy on drive 00 of the KITTI  dataset. Over these 90 runs, the \textit{keypoint} and \textit{L2} strategies respectively detected 292 and 14 false localizations while \textit{RF\_eigen} and \textit{RF\_eigen+shapes} made no false detections.}
\label{fig_relocalization_metric}
\vspace{-5mm}
\end{figure}

In order to show the reproducibility of the results and because the computer load affects the locations at which localization is performed in each run, we perform 90 runs for each strategy and present the average results.
The distance travelled between each localization is recorded and evaluated in a similar manner to \cite{linegar2015work}.
Fig.~\ref{fig_relocalization_metric} shows the probability of travelling a given distance without successful localization in the target map.
Specifically, this metric is computed as follows.
\begin{equation}
\label{localization_metric}
P(x) = \frac{
\begin{matrix}
\text{Sum of distance travelled without} \\
\text{localization for greater or equal to } x \text{ meters}
\end{matrix}}
{\text{Total distance travelled}}
\end{equation}

\vspace{10px}

Although \textit{RF\_eigen+shapes} is the most complex and computationally demanding strategy (see Table~\ref{tab_timing}), it never required more than 55 meters before successful localization, as compared to 67 and 88 meters for \textit{L2} and \textit{RF\_eigen} respectively.
On the other hand, while \textit{L2} is the quickest strategy, it also made 14 false localizations, which could motivate further reduction of the operating point of 0.2 FPR.
For the two random forest based strategies, the vehicle can successfully localize within 35 meters 95\% of the time.
%

Finally, all segment matching methods clearly outperform the keypoint baseline which necessitated much more work to deliver positive results.
Based on keypoint matching, we were not able to obtain an interesting number of true positives without allowing for some false positives.
That is, on average over a one minute localization run, the baseline detected 5.23 true positive and 3.25 false positive localizations.

The computational requirements of this algorithm on an Intel i7-4900MQ CPU @ 2.80GHz are depicted in Table~\ref{tab_timing}
%
%
Note that all operations including ensemble of shape feature extraction, histogram intersection, and random forest classification could benefit of parallelization.





\begin{table}
  \renewcommand{\arraystretch}{1.3}
  \centering
  \caption{Timing of each of the localization modules (in ms).}
  \label{tab_timing}
  \resizebox{3.3in}{!}{%
  \begin{tabular}{|c|c|c|c|}
    \hline
    Module & \textit{L2} & \textit{RF\_eigen}  & \textit{RF\_eigen+shapes} \\
    \hline
    Segmentation & 428.80 $\pm$ 5.83 & 428.78 $\pm$ 6.28 & 435.56 $\pm$ 7.34 \\
Description & 1.37 $\pm$ 0.03 & 1.37 $\pm$ 0.03 & 103.84 $\pm$ 2.41 \\
Matching & 244.52 $\pm$ 9.76 & 289.75 $\pm$ 10.56 & 563.23 $\pm$ 11.96 \\
Geometric verification & 67.43 $\pm$ 2.77 & 76.00 $\pm$ 2.75 & 85.57 $\pm$ 3.16 \\
\hline
Total & 742.12 & 795.91 & 1188.20 \\
    \hline
  \end{tabular}}
  \vspace{-4mm}
\end{table}

\subsection{Loop-closure performance}
\label{ssec:loop_closure_performance}

We now show how our segment based loop detection algorithm can be used online and how it can easily be integrated with a pose-graph trajectory estimation system.
In this scenario, the target map is built online by accumulating segments extracted from the source clouds, as opposed to being loaded before the experiment as performed in Section~\ref{ssec:localization_performance}.
Special care is taken to avoid cluttering the target map with `duplicate segments', i.e. segments resulting from the same object part, but segmented at different times and overlapping in the target map.

\begin{figure*}[!t]
\centering
\subfloat[]{
\includegraphics[width=2.0in]{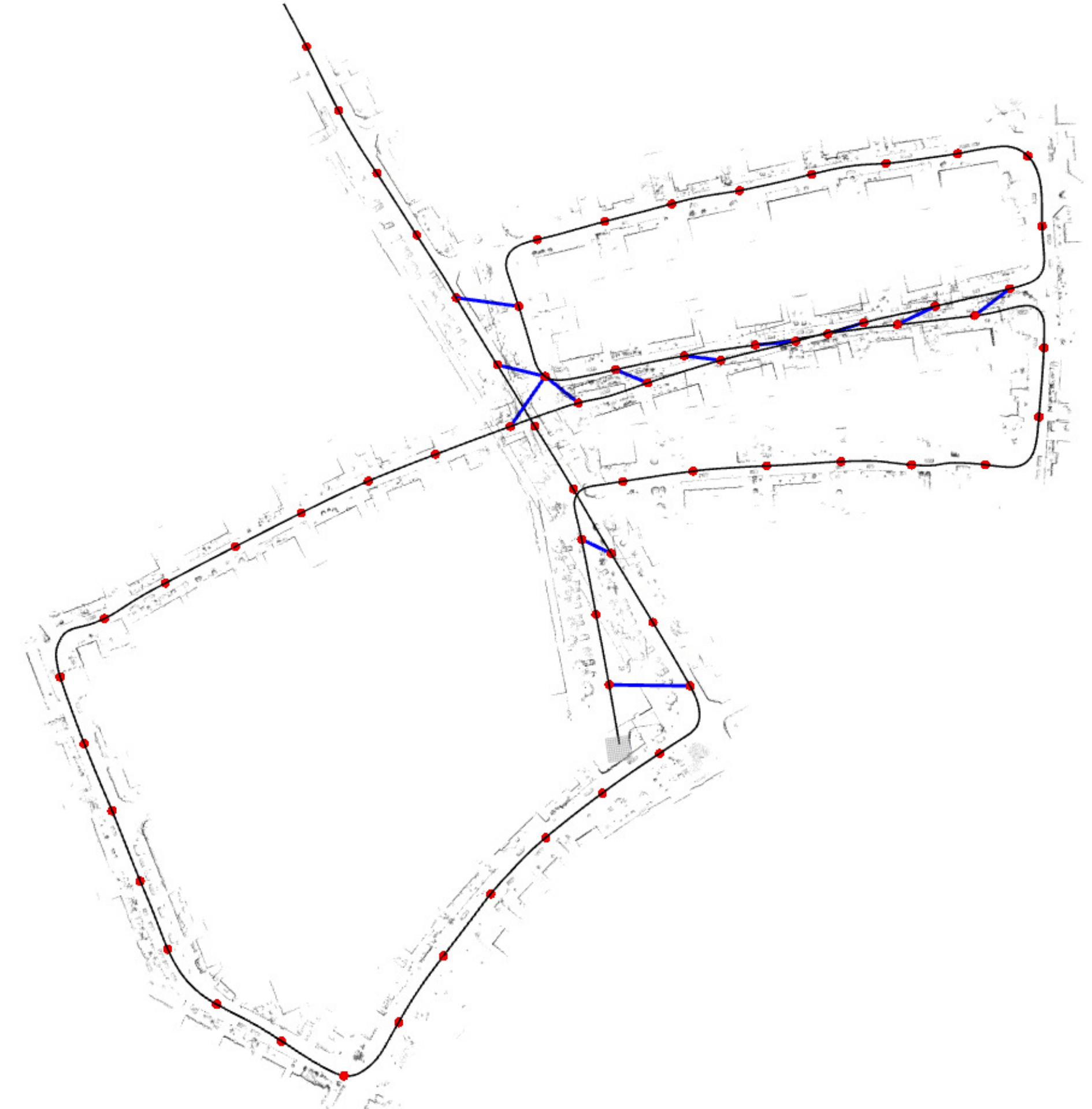}
\label{fig_online_before_optimization}}
\hfil
\subfloat[]{
\includegraphics[width=2.3in]{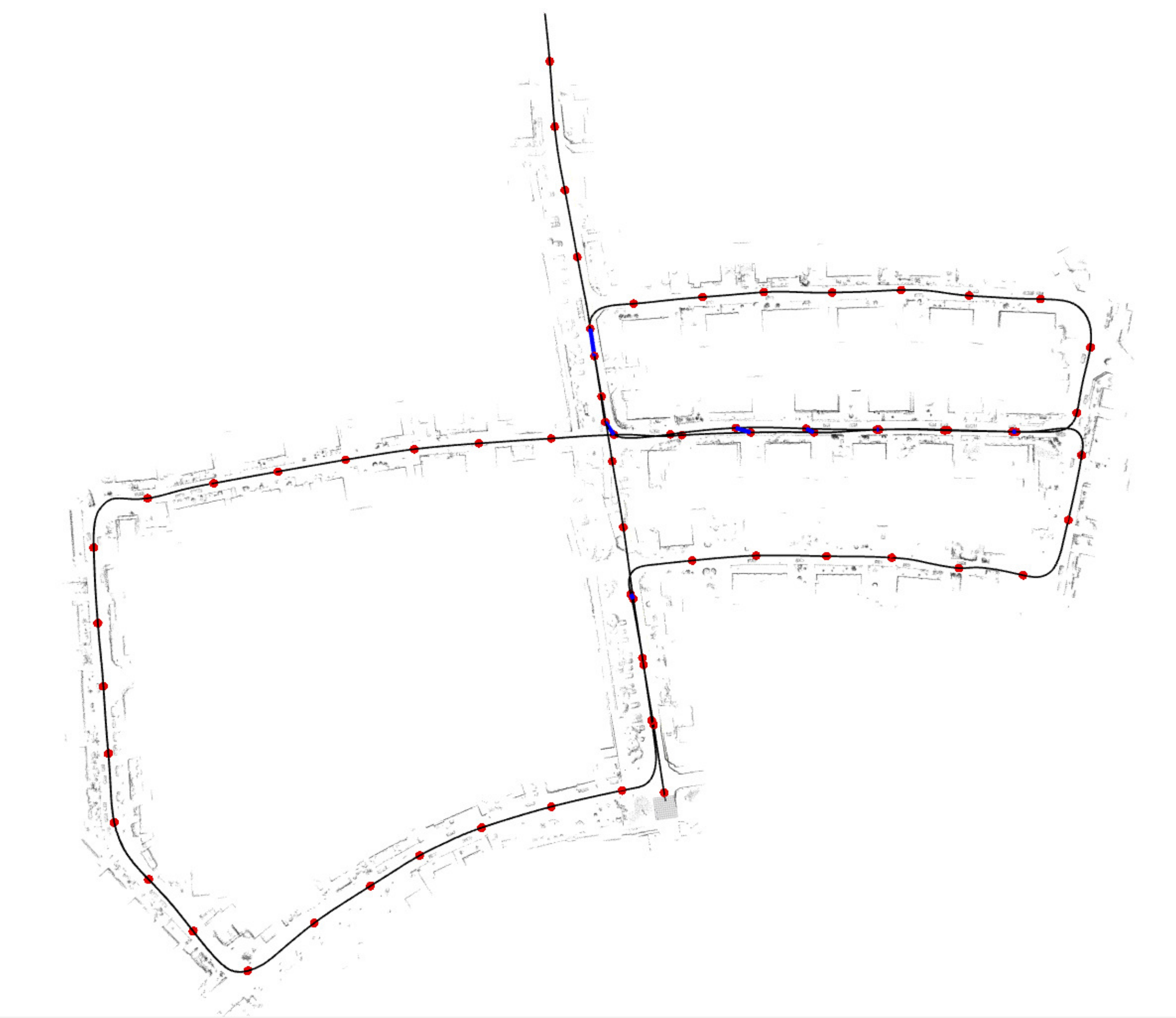}
\label{fig_online_after_optimization}}
\caption{Illustration of loop-closure with \textit{SegMatch}: (a) Shows loops detected in real time by the segment based strategy \textit{RF\_eigen+shapes} during sequence 05 of the KITTI dataset. The red dots represent locations where segmentation and loop-closure detection were performed and the blue lines indicate the detected loops (with no false positives). For illustration purposes, we show the odometry trajectory in (a) whereas (b) illustrates the result of feeding these loops to an online pose-graph trajectory estimator.
}
\label{fig_online_before_optimization}
\vspace{-5mm}
\end{figure*}

The results of applying this strategy on sequence 05 of the KITTI dataset is illustrated in Fig.~\ref{fig_online_before_optimization}.
For this sequence, the global map is created using \ac{ICP} for adding constraints between Velodyne scans.
This introduces a drift over time, as expected in GPS-free state estimation solutions.
On this sequence, our real-time algorithm successfully detected 12 true positive and no false positive loop-closures.
Once loops are detected, they are fed in a pose-graph optimization system similar to the one described in \cite{dube2016non}\footnote{This separate contribution is available at \url{https://github.com/ethz-asl/laser_slam}.}.
The result of this optimization is used to update the target segment positions and remove duplicate segments from the target map as aforementioned.

\subsection{Demonstration with more complex data}
\label{ssec:clausius}

\begin{figure}
\centering
\includegraphics[width=3.4in]{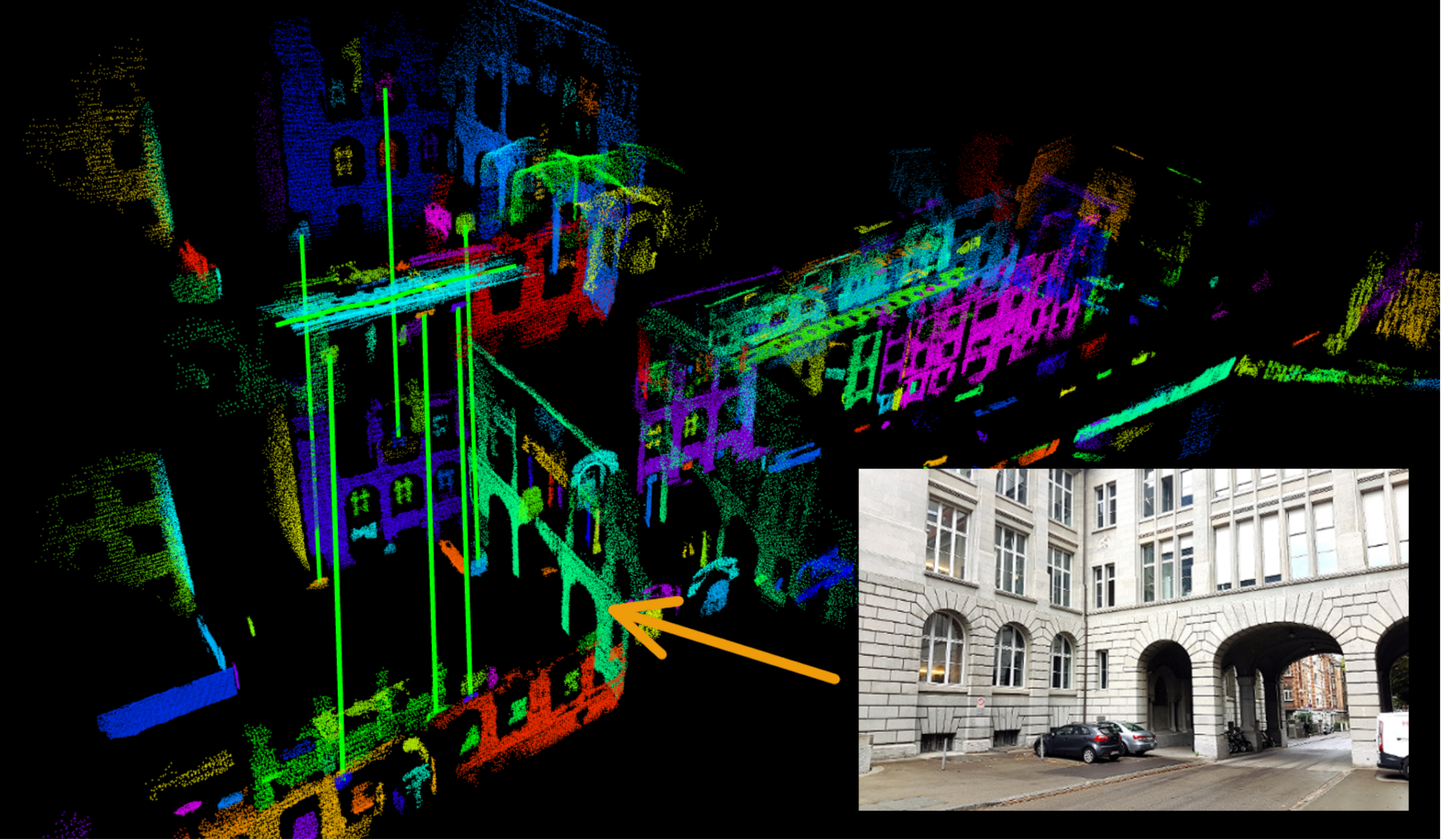}
\caption{An illustration of a correct loop-closure detection based on region growing segmentation with smoothness constraints on data from the Clausiusstrasse in Zurich. The reference point cloud is shown below, and the local point cloud is aligned above. Colours are used to show the point cloud segmentation, and segment matches are indicated with green lines. Note the parts of larger structures (windows, arcs, façades).}
\label{fig_second_segmentation}
\vspace{-5mm}
\end{figure}

To conclude the experiment section, we briefly show that the proposed place recognition algorithm can be applied to other environments and sensor modalities by simply replacing sub-modules of the pipeline.
As an example, in situations where the segmentation algorithm described in Section~\ref{sec_feature_extraction} cannot be applied, this module can be replaced by a different algorithm.
Fig.~\ref{fig_second_segmentation} shows an example of a correct loop detection by matching segments obtained from segmenting the point cloud based on region growing with smoothness constraints \cite{rabbani2006segmentation}.
Although these types of segments may appear to be less meaningful for humans, they provide discriminative features for the loop-closure algorithm, as illustrated by the matches shown in Fig.~\ref{fig_second_segmentation}.

\section{CONCLUSION}
\label{sec:conclusion}

This paper presented \textit{SegMatch}, an algorithm for performing place recognition in 3D laser data based on the concept of segment matching.
Compared to a keypoint approach, acting at the level of segments offers several advantages without making any assumptions about perfect segmentation or on the presence of `objects' in the environment.
Our modular approach first extracts segments from a source point cloud, which are then described and matched to previously mapped target segments.
A geometric-verification step is finally applied to turn these candidate matches into place recognition candidates.

This framework has been exhaustively evaluated on the KITTI dataset.
We first analysed the impact of using a random forest classifier to learn an adequate feature distance metric for the purpose of matching segments.
We have then shown that the algorithm is able to accurately localize at a frequency higher than 1Hz in the largest map of the KITTI dataset.
We also demonstrated how it is possible to robustly detect loops in an online fashion, and how these can be fed to a pose-graph trajectory estimator.
Thanks to the framework's modular approach, we have furthermore illustrated that it can easily be applied to different scenarios by simply changing building blocks of the algorithm.
The source code for the entire framework is available online, offering real-time segmentation and loop-closure detection for streams of 3D point clouds.

Based on this segment matching technique, we foresee several possible advantages in systems which do more than mapping - using segments for both matching and describing the environment.
We will pursue supervised learning techniques to interpret these \emph{segment}-based maps into structural and object semantic classes.









\begin{acronym}
\acro{ICP}{Iterative Closest Point}
\acro{MAP}{Maximum A Posteriori}
\acro{SLAM}{Simultaneous Localization and Mapping}
\acro{DOF}{Degrees of Freedom}
\acro{GUI}{Graphical User Interface}
\acro{TRADR}{``Long-Term Human-Robot Teaming for Robots Assisted Disaster Response''}
\acro{k-NN}{k-Nearest Neighbors}
\end{acronym}

\small
\bibliographystyle{IEEEtranN}
\bibliography{IEEEabrv,icra2017}

\begin{thebibliography}{30}
\providecommand{\natexlab}[1]{#1}
\providecommand{\url}[1]{#1}
\csname url@samestyle\endcsname
\providecommand{\newblock}{\relax}
\providecommand{\bibinfo}[2]{#2}
\providecommand{\BIBentrySTDinterwordspacing}{\spaceskip=0pt\relax}
\providecommand{\BIBentryALTinterwordstretchfactor}{4}
\providecommand{\BIBentryALTinterwordspacing}{\spaceskip=\fontdimen2\font plus
\BIBentryALTinterwordstretchfactor\fontdimen3\font minus
  \fontdimen4\font\relax}
\providecommand{\BIBforeignlanguage}[2]{{%
\expandafter\ifx\csname l@#1\endcsname\relax
\typeout{** WARNING: IEEEtranN.bst: No hyphenation pattern has been}%
\typeout{** loaded for the language `#1'. Using the pattern for}%
\typeout{** the default language instead.}%
\else
\language=\csname l@#1\endcsname
\fi
#2}}
\providecommand{\BIBdecl}{\relax}
\BIBdecl

\bibitem[Thrun et~al.(2002)]{thrun2002robotic}
S.~Thrun \emph{et~al.}, ``Robotic mapping: A survey,'' \emph{Exploring
  artificial intelligence in the new millennium}, vol.~1, pp. 1--35, 2002.

\bibitem[Lowry et~al.(2016)Lowry, Sunderhauf, Newman, Leonard, Cox, Corke, and
  Milford]{lowryvisual}
S.~Lowry, N.~Sunderhauf, P.~Newman, J.~J. Leonard, D.~Cox, P.~Corke, and M.~J.
  Milford, ``Visual place recognition: A survey,'' \emph{{IEEE} Trans. on
  Robotics}, 2016.

\bibitem[Bosse and Zlot(2013)]{bosse2013place}
M.~Bosse and R.~Zlot, ``Place recognition using keypoint voting in large {3D}
  lidar datasets,'' in \emph{{IEEE} Int. Conf. on Robotics and Automation},
  2013.

\bibitem[Alexe et~al.(2010)Alexe, Deselaers, and Ferrari]{alexe2010cvpr}
B.~Alexe, T.~Deselaers, and V.~Ferrari, ``What is an object?'' in \emph{{IEEE}
  Conf. on Computer Vision and Pattern Recognition}, 2010.

\bibitem[Scovanner et~al.(2007)Scovanner, Ali, and Shah]{scovanner20073}
P.~Scovanner, S.~Ali, and M.~Shah, ``A 3-dimensional sift descriptor and its
  application to action recognition,'' in \emph{{ACM} Int. Conf. on
  Multimedia}, 2007.

\bibitem[Wohlkinger and Vincze(2011)]{wohlkinger2011ensemble}
W.~Wohlkinger and M.~Vincze, ``Ensemble of shape functions for 3d object
  classification,'' in \emph{{IEEE} Int. Conf. on Robotics and Biomimetics},
  2011.

\bibitem[Rusu et~al.(2009)Rusu, Blodow, and Beetz]{rusu2009fast}
R.~B. Rusu, N.~Blodow, and M.~Beetz, ``Fast point feature histograms (fpfh) for
  3d registration,'' in \emph{{IEEE} Int. Conf. on Robotics and Automation},
  2009, pp. 3212--3217.

\bibitem[Zhuang et~al.(2013)Zhuang, Jiang, Hu, and Yan]{zhuang20133}
Y.~Zhuang, N.~Jiang, H.~Hu, and F.~Yan, ``3-d-laser-based scene measurement and
  place recognition for mobile robots in dynamic indoor environments,''
  \emph{{IEEE} Transactions on Instrumentation and Measurement}, vol.~62,
  no.~2, pp. 438--450, 2013.

\bibitem[Steder et~al.(2010)Steder, Grisetti, and Burgard]{steder2010robust}
B.~Steder, G.~Grisetti, and W.~Burgard, ``Robust place recognition for {3D}
  range data based on point features,'' in \emph{{IEEE} Int. Conf. on Robotics
  and Automation}, 2010.

\bibitem[Steder et~al.(2011)Steder, Ruhnke, Grzonka, and
  Burgard]{steder2011place}
B.~Steder, M.~Ruhnke, S.~Grzonka, and W.~Burgard, ``Place recognition in 3d
  scans using a combination of bag of words and point feature based relative
  pose estimation,'' in \emph{{IEEE/RSJ} Int. Conf. on Intelligent Robots and
  Systems}, 2011.

\bibitem[Gawel et~al.(2016)Gawel, Cieslewski, Dub{\'{e}}, Bosse, Siegwart, and
  Nieto]{Gawel2016}
A.~Gawel, T.~Cieslewski, R.~Dub{\'{e}}, M.~Bosse, R.~Siegwart, and J.~Nieto,
  ``{Structure-based Vision-Laser Matching},'' in \emph{{IEEE/RSJ} Int. Conf.
  on Intelligent Robots and Systems}, Daejeon, 2016.

\bibitem[Zhang and Singh(2014)]{zhang2014loam}
J.~Zhang and S.~Singh, ``Loam: Lidar odometry and mapping in real-time,'' in
  \emph{Robotics: Science and Systems}, 2014.

\bibitem[Rohling et~al.(2015)Rohling, Mack, and Schulz]{rohling2015fast}
T.~Rohling, J.~Mack, and D.~Schulz, ``A fast histogram-based similarity measure
  for detecting loop closures in 3-d lidar data,'' in \emph{{IEEE/RSJ} Int.
  Conf. on Intelligent Robots and Systems}, 2015.

\bibitem[Granstr{\"o}m et~al.(2011)Granstr{\"o}m, Sch{\"o}n, Nieto, and
  Ramos]{granstrom2011learning}
K.~Granstr{\"o}m, T.~B. Sch{\"o}n, J.~I. Nieto, and F.~T. Ramos, ``Learning to
  close loops from range data,'' \emph{The Int. Journal of Robotics Research},
  vol.~30, no.~14, pp. 1728--1754, 2011.

\bibitem[Magnusson et~al.(2009)Magnusson, Andreasson, N{\"u}chter, and
  Lilienthal]{magnusson2009automatic}
M.~Magnusson, H.~Andreasson, A.~N{\"u}chter, and A.~J. Lilienthal, ``Automatic
  appearance-based loop detection from three-dimensional laser data using the
  normal distributions transform,'' \emph{Journal of Field Robotics}, vol.~26,
  no. 11-12, pp. 892--914, 2009.

\bibitem[Fernandez-Moral et~al.(2013)Fernandez-Moral, Mayol-Cuevas, Arevalo,
  and Gonzalez-Jimenez]{fernandez2013fast}
E.~Fernandez-Moral, W.~Mayol-Cuevas, V.~Arevalo, and J.~Gonzalez-Jimenez,
  ``Fast place recognition with plane-based maps,'' in \emph{{IEEE} Int. Conf.
  on Robotics and Automation}, 2013.

\bibitem[Fern{\'a}ndez-Moral et~al.(2016)Fern{\'a}ndez-Moral, Rives,
  Ar{\'e}valo, and Gonz{\'a}lez-Jim{\'e}nez]{fernandez2016scene}
E.~Fern{\'a}ndez-Moral, P.~Rives, V.~Ar{\'e}valo, and
  J.~Gonz{\'a}lez-Jim{\'e}nez, ``Scene structure registration for localization
  and mapping,'' \emph{Robotics and Autonomous Systems}, vol.~75, pp. 649--660,
  2016.

\bibitem[Finman et~al.(2015)Finman, Paull, and Leonard]{finman2015icraws}
R.~Finman, L.~Paull, and J.~J. Leonard, ``Toward object-based place recognition
  in dense rgb-d maps,'' in \emph{ICRA workshop on visual place recognition in
  changing environments}, 2015.

\bibitem[Douillard et~al.(2012)Douillard, Quadros, Morton, Underwood, De~Deuge,
  Hugosson, Hallstr{\"o}m, and Bailey]{douillard2012scan}
B.~Douillard, A.~Quadros, P.~Morton, J.~P. Underwood, M.~De~Deuge, S.~Hugosson,
  M.~Hallstr{\"o}m, and T.~Bailey, ``Scan segments matching for pairwise 3d
  alignment,'' in \emph{{IEEE} Int. Conf. on Robotics and Automation}, 2012.

\bibitem[Nieto et~al.(2006)Nieto, Bailey, and Nebot]{nieto2006scan}
J.~Nieto, T.~Bailey, and E.~Nebot, ``Scan-slam: Combining ekf-slam and scan
  correlation.''\hskip 1em plus 0.5em minus 0.4em\relax Springer, 2006, pp.
  167--178.

\bibitem[Douillard et~al.(2014)Douillard, Underwood, Vlaskine, Quadros, and
  Singh]{douillard2014pipeline}
B.~Douillard, J.~Underwood, V.~Vlaskine, A.~Quadros, and S.~Singh, ``A pipeline
  for the segmentation and classification of 3d point clouds,'' in
  \emph{Experimental Robotics}.\hskip 1em plus 0.5em minus 0.4em\relax
  Springer, 2014, pp. 585--600.

\bibitem[Douillard et~al.(2011)Douillard, Underwood, Kuntz, Vlaskine, Quadros,
  Morton, and Frenkel]{douillard2011segmentation}
B.~Douillard, J.~Underwood, N.~Kuntz, V.~Vlaskine, A.~Quadros, P.~Morton, and
  A.~Frenkel, ``On the segmentation of 3d lidar point clouds,'' in \emph{{IEEE}
  Int. Conf. on Robotics and Automation}, 2011.

\bibitem[Weinmann et~al.(2014)Weinmann, Jutzi, and
  Mallet]{weinmann2014semantic}
M.~Weinmann, B.~Jutzi, and C.~Mallet, ``Semantic 3d scene interpretation: a
  framework combining optimal neighborhood size selection with relevant
  features,'' \emph{ISPRS Annals of the Photogrammetry, Remote Sensing and
  Spatial Information Sciences}, vol.~2, no.~3, p. 181, 2014.

\bibitem[Breiman(2001)]{breiman2001random}
L.~Breiman, ``Random forests,'' \emph{Machine Learning}, vol.~45, no.~1, pp.
  5--32, 2001.

\bibitem[Fischler and Bolles(1981)]{fischler1981random}
M.~A. Fischler and R.~C. Bolles, ``Random sample consensus: a paradigm for
  model fitting with applications to image analysis and automated
  cartography,'' \emph{Communications of the ACM}, vol.~24, no.~6, pp.
  381--395, 1981.

\bibitem[Geiger et~al.(2012)Geiger, Lenz, and Urtasun]{geiger2012we}
A.~Geiger, P.~Lenz, and R.~Urtasun, ``Are we ready for autonomous driving? the
  kitti vision benchmark suite,'' in \emph{{IEEE} Conf. on Computer Vision and
  Pattern Recognition}, 2012.

\bibitem[Rusu and Cousins(2011)]{Rusu_ICRA2011_PCL}
R.~B. Rusu and S.~Cousins, ``{3D is here: Point Cloud Library (PCL)},'' in
  \emph{{IEEE} Int. Conf. on Robotics and Automation}, 2011.

\bibitem[Linegar et~al.(2015)Linegar, Churchill, and Newman]{linegar2015work}
C.~Linegar, W.~Churchill, and P.~Newman, ``Work smart, not hard: Recalling
  relevant experiences for vast-scale but time-constrained localisation,'' in
  \emph{{IEEE} Int. Conf. on Robotics and Automation}, 2015.

\bibitem[Dubé et~al.(2016)Dubé, Sommer, Gawel, Bosse, and
  Siegwart]{dube2016non}
R.~Dubé, H.~Sommer, A.~Gawel, M.~Bosse, and R.~Siegwart, ``Non-uniform
  sampling strategies for continuous correction based trajectory estimation,''
  in \emph{{IEEE} Int. Conf. on Robotics and Automation}, 2016.

\bibitem[Rabbani et~al.(2006)Rabbani, Van Den~Heuvel, and
  Vosselmann]{rabbani2006segmentation}
T.~Rabbani, F.~Van Den~Heuvel, and G.~Vosselmann, ``Segmentation of point
  clouds using smoothness constraint,'' \emph{International Archives of
  Photogrammetry, Remote Sensing and Spatial Information Sciences}, vol.~36,
  no.~5, pp. 248--253, 2006.

\end{thebibliography}

\end{document}